\documentclass[letterpaper, 10pt, conference]{ieeeconf}

\IEEEoverridecommandlockouts
\usepackage{amsmath,amssymb,amsfonts}
\usepackage{graphicx}
\usepackage{booktabs}
\usepackage{cite}
\usepackage{float}
\usepackage{bm}
\usepackage{siunitx}
\usepackage{url}

\newcommand{\R}{\mathbb{R}}
\newcommand{\state}{\bm{x}}
\newcommand{\meas}{\bm{z}}
\newcommand{\mat}[1]{\mathbf{#1}}
\newcommand{\parhead}[1]{\noindent\hspace*{\parindent}\textbf{#1}\quad\ignorespaces}

\title{\LARGE \bf
TinyCVIO: A Constellation-Aided Visual-Inertial Odometry \\ System for Nanodrones
}

\author{%
  Derin Ozturk, Kaan Akan, Irwin Wang, and Christopher Batten\thanks{All authors are with the Department of ECE, Cornell University, Ithaca, NY, USA. Corresponding author: {\tt\small ddo26@cornell.edu}. Video: \protect\url{https://youtu.be/6hQNdIcjHJE}.}%
}

\begin{document}

\bstctlcite{IEEErefcontrol}

\maketitle
\thispagestyle{empty}
\pagestyle{empty}

\begin{abstract}
Nanodrones require accurate, real-time state estimation under severe sensing and computational constraints.
We present TinyCVIO, a visual-inertial odometry system that co-designs miniature sensing, visual processing, and estimation for a commodity dual-core microcontroller with \SI{520}{\kilo\byte} SRAM.
Lightweight LED constellations provide known geometry without surveyed positions or yaw angles, assuming placement on a common level plane.
A streaming visual frontend tracks LED observations from a millimeter-scale camera at \SI{29.2}{FPS}, while a rigid-board measurement model retains inter-LED constraints and streaming QR bounds estimation workspace for a fixed filter-state size.
Across 19 hand-held hardware-in-the-loop datasets, the rigid-board model reduces mean absolute trajectory error by 27\% relative to planar points.
The complete system runs onboard a Crazyflie across nine flights at three speeds, achieving \SIrange{3.5}{3.7}{\centi\meter} mean absolute trajectory error and 0.50--0.60\% relative pose error over \SI{10}{\meter} segments, with mean estimate latency of \SIrange{15.7}{16.3}{\milli\second}.
\end{abstract}

\section{INTRODUCTION}
Nanodrones are compelling platforms for exploration, inspection, and swarm deployment, but their size, weight, and power budgets force a tradeoff between sensor fidelity and onboard compute.
State-estimation strategies based on minimal optical-flow and inertial sensing~\cite{honeggerOpenSourceOpen2013,yuTinySenseLighterWeight2025} support hover within these budgets but lack an absolute horizontal position reference, allowing position drift to accumulate.
Improving estimation capability typically introduces additional complexity, either through richer onboard sensing and computation or by instrumenting the environment.
Approaches that increase onboard capability have used microcontrollers with larger compute and memory budgets~\cite{gangstadBareMetalVisualInertial2024}, specialized multicore companion processors~\cite{niculescuNanoSLAMEnablingFully2024,palossiOpenSourceOpen2019}, or custom silicon~\cite{suleimanNavion2mWFully2019}; others offload processing to a ground station~\cite{zhengMonocularDepthEstimation2024}.
Externally referenced systems such as motion capture, Lighthouse, and ultra-wideband (UWB) instead shift requirements toward environmental infrastructure, where equipment cost, installation, and spatial calibration complicate deployment and expansion~\cite{taffanelLighthousePositioningSystem2021,grassoAnalysisAccuracyImprovement2022,zhaoLearningBasedBiasCorrection2021}.
We seek to improve estimation capability while keeping onboard resource demands, system complexity, and deployment effort low.

\begin{figure}[t]
  \centering
  \includegraphics[width=0.99\columnwidth]{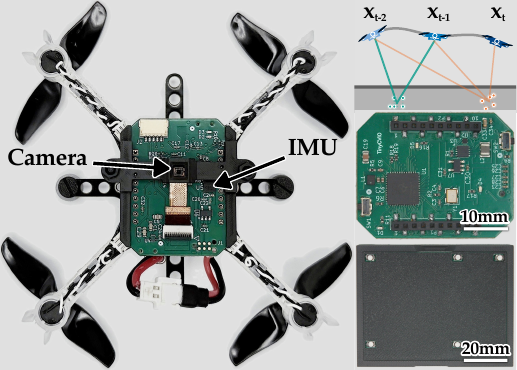}
  \caption{TinyCVIO hardware and estimation concept.
    \emph{Left:} Crazyflie carrying our custom TinyCVIO deck, with the NanEyeC camera connected through a custom flexible printed circuit board cable and the ICM-45686 IMU indicated.
    \emph{Right, top to bottom:} observations of shared LED constellations across successive camera poses; RP2354-based companion compute deck for a Crazyflie; and five-LED constellation with known geometry.}
  \label{fig:overview}
\end{figure}

We present TinyCVIO, a visual-inertial odometry system that jointly designs the real-time implementation, sensing strategy, and estimation algorithm around scalable deployments of low-cost, battery-powered LED constellations and a commodity RP2354 dual-core microcontroller.
Each constellation carries five LEDs in a known pattern; board positions and yaw angles need not be surveyed, assuming placement on a common level plane.
This lightweight instrumentation provides both visually distinctive landmarks for a millimeter-scale camera and known geometric relationships that the estimator can exploit.
The visual frontend runs bit-serial preprocessing concurrently with geometry-based constellation detection and tracking on one core.
On the other core, the backend exploits the constellations' rigid geometry in its measurement model and uses streaming computation to bound estimation workspace.
We demonstrate the complete pipeline running onboard a Crazyflie in real time during flight.

Our contributions are (1) a system for real-time onboard nanodrone VIO using commodity compute and miniature sensing; (2) a visual frontend combining CPU-independent bit-serial preprocessing with geometry-based constellation detection and tracking; (3) a rigid-board MSCKF measurement model and streaming estimation strategy that preserve geometric constraints under tight memory budgets; and (4) evaluation through simulation, live hardware-in-the-loop experiments, paired replays, and onboard flight.
The model extends to other rigid fiducials under the same placement assumptions, and the streaming strategy applies to other memory-constrained MSCKF systems.
We will open-source the hardware and software to support nanodrone research.

\section{Related Work}
\label{sec:related}

We survey related work across three areas: localization strategies for nanodrones, structured landmarks and pose estimation, and embedded visual-inertial odometry.

\subsection{Localization and Onboard Sensing}
\label{sec:related_loc}

Infrastructure-based localization supports Crazyflie-class platforms but requires site instrumentation.
Lighthouse relies on line-of-sight to dedicated base stations~\cite{taffanelLighthousePositioningSystem2021}, while UWB accuracy varies with deployment geometry and measurement bias~\cite{grassoAnalysisAccuracyImprovement2022,zhaoLearningBasedBiasCorrection2021}.
Optical-flow--inertial systems support hover with minimal onboard resources~\cite{honeggerOpenSourceOpen2013,yuTinySenseLighterWeight2025}, but lack an absolute horizontal position reference.

Richer onboard perception has used laptop offloading for monocular depth-based navigation~\cite{zhengMonocularDepthEstimation2024}, GAP8/GAP9 companion processors for reactive obstacle avoidance or 2D SLAM~\cite{palossiOpenSourceOpen2019,niculescuNanoSLAMEnablingFully2024}, and custom silicon for VIO~\cite{suleimanNavion2mWFully2019}.

\subsection{Structured Landmarks and Pose Estimation}
\label{sec:related_marker}

Active LED markers provide robust detection under challenging lighting and clutter.
An external-camera approach~\cite{faesslerMonocularPoseEstimation2014} uses asymmetrically placed IR LEDs, solving correspondences via combinatorial P3P and achieving sub-centimeter accuracy, but processing runs on a laptop with no IMU fusion.
Our prior MCU-class system~\cite{ozturkAbsolutePoseEstimation2024} demonstrates absolute pose estimation on a Cortex-M4 using a millimeter-scale camera and planar LED landmarks with known positions and orientations, reaching 16.5\,FPS at ${\sim}\SI{15}{\milli\meter}$ accuracy without temporal filtering.

Unlike these pose-only approaches, TinyCVIO fuses visual and inertial measurements over time.

OV\_plane~\cite{chenMonocularVisualInertialOdometry2023} exploits planar structure by constraining natural point features to planes maintained in the filter state.
TinyCVIO instead exploits known within-constellation geometry and eliminates the shared board pose through nullspace projection.

\subsection{Embedded Visual-Inertial Odometry}
\label{sec:related_vio}

OpenVINS~\cite{genevaOpenVINSResearchPlatform2020} provides an open-source implementation of MSCKF-based visual-inertial estimation~\cite{mourikisMultiStateConstraintKalman2007}, but is not directly suitable for bare-metal deployment within our microcontroller's memory and compute constraints.
A bare-metal VIO implementation~\cite{gangstadBareMetalVisualInertial2024} demonstrates execution on a Cortex-M7 at \SI{1}{\giga\hertz} with \SI{3}{\mega\byte} RAM, establishing MCU feasibility with substantially greater memory resources than TinyCVIO.

Recent GAP9 implementations include downfacing VIO with a four-DOF motion model and a floor-range sensor~\cite{kuehneEfficientAccurateDownfacing2025}, and LEVIO for full six-DOF visual-inertial estimation~\cite{kuehneLEVIO2026}.
Their evaluations combine recorded sensor data with embedded profiling; neither reports the complete pipeline operating onboard a nanodrone in flight.

\begin{figure}[!b]
  \centering
  \includegraphics[width=0.99\columnwidth]{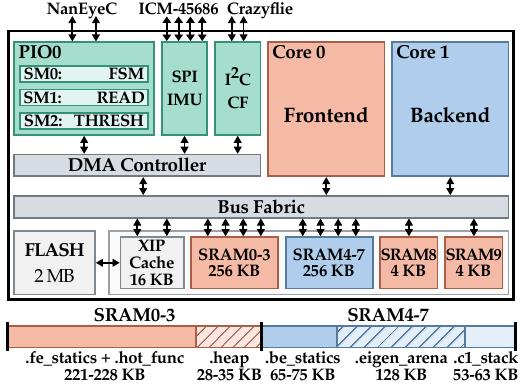}
  \caption{RP2354 architecture and reserved memory allocation. The lower map omits Core~0 stack allocation in SRAM8--9.}
  \label{fig:sys_arch}
\end{figure}

\section{System Design}
\label{sec:system}

TinyCVIO combines a monocular camera and IMU with dual-core processing on a custom nanodrone PCB.
Programmable I/O offloads camera readout and thresholding, while separate cores execute the visual frontend (Sec.~\ref{sec:frontend}) and visual-inertial backend (Sec.~\ref{sec:backend}), respectively (Fig.~\ref{fig:sys_arch}).

\subsection{Hardware and Constellations}
\label{sec:hardware}
\label{sec:constellation}

TinyCVIO uses an RP2354 dual-core Cortex-M33 microcontroller clocked at \SI{300}{\mega\hertz}, with \SI{520}{\kilo\byte} SRAM and \SI{2}{\mega\byte} on-chip flash~\cite{rp2350datasheet}.
A monocular NanEyeC camera provides $320 \times 320$ images, and an ICM-45686 IMU supplies accelerometer and gyroscope measurements at \SI{400}{\hertz}.
The camera operates at \SI{29.2}{FPS}, giving approximately \SI{34.2}{\milli\second} between frames.
The complete deck as pictured in Fig.~\ref{fig:overview}, including the 3D-printed camera shroud, weighs \SI{5.1}{\gram}.

Each battery-powered constellation carries four LEDs at the corners of a \SI{40}{\milli\meter} square and a fifth disambiguator \SI{19}{\milli\meter} from one corner, breaking rotational symmetry.
Boards lie on a common level plane, with unsurveyed positions and yaw angles.
The known LED geometry provides the rigid constraints used by the estimator.

\subsection{Streaming Sensor Acquisition}
\label{sec:acquisition}

Programmable I/O (PIO) provides hardware state machines that execute a compact assembly instruction set independently of the CPU.
Our camera framing, pixel-extraction, and thresholding programs together occupy all 32 instruction slots shared by three state machines in one PIO block.
The pipeline extracts 8-bit pixel values from the camera's 12-bit serial packets and generates threshold events for the visual frontend, with exposure and threshold set so that events are dominated by direct LED light.
A statically reserved half-image buffer supports DMA image streaming during hardware-in-the-loop (HITL) experiments.
Onboard estimation does not require this buffer.
A FIFO watermark interrupt triggers IMU readout over SPI on Core~0, which enqueues timestamped inertial samples for propagation on Core~1.

\subsection{Concurrent Execution}
\label{sec:firmware}

Core~0 performs sensor acquisition and visual tracking, while Core~1 performs inertial propagation and visual-inertial updates.
Lock-free buffers transfer visual observations and IMU samples.
Core~1 propagates the state as each watermark-triggered IMU batch becomes available, spreading propagation work across the inter-frame interval rather than accumulating it before a visual update.
The estimator consumes the latest available visual frame without blocking the frontend.
The frontend's live path uses static and stack allocation; backend matrices use a fixed SRAM arena.
Figure~\ref{fig:sys_arch} shows reserved capacities; ranges reflect different evaluated configurations.
Pose estimates are sent to the Crazyflie flight controller over I\textsuperscript{2}C.

\section{Visual Frontend}
\label{sec:frontend}

\begin{figure}[!b]
  \centering
  \includegraphics[width=0.99\columnwidth]{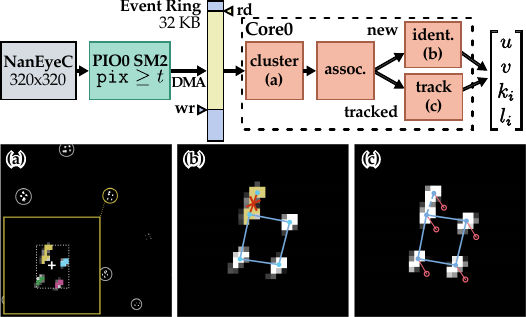}
  \caption{Visual frontend pipeline and example observations.
    (a) Spatial grouping and connected components, with an enlarged constellation.
    (b) Geometry-based LED identification, including candidate separation of a merged component.
    (c) Tracking preserves LED identities between frames using a fitted similarity transform; open circles denote previous positions and filled circles current tracked positions.
    Outputs contain image coordinates and board and LED identifiers.}
  \label{fig:frontend_dataflow}
\end{figure}

The frontend converts threshold events into labeled LED observations for visual-inertial estimation (Fig.~\ref{fig:frontend_dataflow}).
Known constellation geometry and temporal tracking maintain correspondences as viewing distance changes: nearby constellations may fragment into separate spatial groups, while individual LED images may merge into a single connected component.

\subsection{Event Clustering}
\label{sec:clustering}

Threshold events arrive in raster order and are processed in a single pass.
The frontend jointly forms spatial groups representing candidate constellations and connected components (CCs) representing contiguous bright pixels within each group.
A component may contain one LED or several merged LEDs.

Each event joins a nearby group or starts a new one.
Grouping radii adapt to the LED spacing observed in previously tracked constellations, accommodating changes in apparent board size.
Within each group, a two-row scanline buffer identifies connected components and accumulates pixel counts and first and second moments.
These statistics support centroid computation and separation of merged LEDs without retaining a full image.

Groups are finalized as readout advances beyond their spatial extent.

\subsection{Association and Identification}
\label{sec:disambiguation}

Candidate constellation groups are matched to existing tracks using spatial proximity, checked against predicted and last-observed positions.
For newly observed constellations, identification assigns each detected LED to its position in the known board geometry.

Identification considers up to 30 LED assignments consistent with the square-plus-disambiguator layout (Sec.~\ref{sec:constellation}).
A closed-form similarity fit ranks these candidates, and IPPE~\cite{collinsInfinitesimalPlaneBasedPose2014} selects the front-facing pose and assignment with the lowest five-point reprojection error.
The similarity fit ranks candidates without rejecting them, since perspective distortion can produce a poor similarity fit even for the correct assignment.

When two LEDs form a single connected component, its accumulated second moments support a principal-component split that generates candidate LED centroids.

\subsection{Tracking and Label Continuity}
\label{sec:sim_tracking}

Once a constellation is identified, tracking maintains its LED correspondences across frames using a closed-form 2D similarity transform (Fig.~\ref{fig:frontend_dataflow}c).
The transform is fitted from previous LED positions to current component centroids; for a merged pair, its previous mean position is matched to the shared centroid.
Isolated LEDs use their measured component centroids, while merged pairs retain positions propagated by the fitted transform.
The resulting configuration is validated against the board geometry and IPPE.
If tracking fails, the frontend attempts nearest-neighbor matching, followed by re-identification with LED labels aligned to the previous ordering.

To suppress persistent label swaps, a continuity check rejects tracked configurations whose square orientation changes by more than a threshold $\theta_{\max}$, set to $30^\circ$ in our experiments, or whose chirality reverses between frames.
Accepted LED coordinates and identifiers are passed to the backend.

\section{Visual-Inertial Backend}
\label{sec:backend}

Our visual-inertial estimation framework is modeled after OpenVINS~\cite{genevaOpenVINSResearchPlatform2020}, with a focus on embedded execution.
Like OpenVINS, it separates the measurement model from propagation, nuisance elimination, measurement compression, and the filter update.
Within the Multi-State Constraint Kalman Filter (MSCKF) framework~\cite{mourikisMultiStateConstraintKalman2007}, we present our streaming QR strategy and rigid-board measurement model, then describe sensor compensation and initialization before comparing independent-point models.

\subsection{State and MSCKF Overview}

The filter state is
\begin{equation}
  \begin{aligned}
    \state &= \begin{bmatrix}
      \state_{\text{IMU}}^\top & t_d & \state_{c_1}^\top & \cdots & \state_{c_N}^\top
    \end{bmatrix}^{\!\top}, \\
    \state_{\text{IMU}} &= \begin{bmatrix}
      {}_G^I\bar{q}^\top & {}^G\mathbf{p}_I^\top & {}^G\mathbf{v}_I^\top & \mathbf{b}_g^\top & \mathbf{b}_a^\top
    \end{bmatrix}^{\!\top},
  \end{aligned}
  \label{eq:state}
\end{equation}
where $\state_{\text{IMU}}$ contains the IMU orientation, position, velocity, and gyroscope/accelerometer biases, and each clone $\state_{c_i}$ stores the camera orientation and position at keyframe $i$.
The IMU and clone error states have dimensions 15 and 6, respectively.
Neither LED positions nor board poses are retained in the filter state.
As in OpenVINS, optional online camera--IMU time-offset calibration adds the scalar state $t_d$ to account for the effective delay introduced by the IMU's onboard low-pass filter during on-drone operation.
IMU propagation, clone management, first-estimates Jacobians, and the standard EKF update follow~\cite{mourikisMultiStateConstraintKalman2007,genevaOpenVINSResearchPlatform2020}.

\parhead{Streaming QR strategy.} \label{sec:streaming_qr}
Our streaming QR strategy performs nuisance elimination within tracks and accumulates the resulting state constraints across tracks without assembling the full measurement stack.
Each track comprises observations of a board or an individual point, depending on the measurement model.
After nuisance elimination and innovation gating, each track's projected Jacobian $\mat{H}_{x,k}^{\perp}$ and residual $\mathbf{r}_k^{\perp}$ are \emph{folded} into a running factor using Givens rotations~\cite{golubMatrixComputations2013}:
\begin{equation}
  \begin{bmatrix}
    \mat{R}_{\text{stream}} & \mathbf{r}_{\text{stream}} \\
    \mat{H}_{x,k}^{\perp} & \mathbf{r}_k^{\perp}
  \end{bmatrix}
  \xrightarrow{\text{Givens}}
  \begin{bmatrix}
    \mat{R}_{\text{new}} & \mathbf{r}_{\text{new}} \\
    \mat{0} & \mathbf{e}_k
  \end{bmatrix}.
  \label{eq:fold}
\end{equation}
The resulting factor and residual replace the running pair.
When rows exceed the factor's column dimension, the zero-Jacobian residual $\mathbf{e}_k$ is discarded.
After processing the accepted tracks, the retained system is
\begin{equation}
  \mathbf{r}_{\text{stream}}
  = \mat{R}_{\text{stream}}\,\delta\state + \tilde{\mathbf{n}}.
  \label{eq:compressed_measurement}
\end{equation}
With columns mapped to the full error state, this system supplies the measurement Jacobian, residual, and transformed noise covariance.
The EKF update and clone marginalization follow the standard MSCKF/OpenVINS procedure~\cite{mourikisMultiStateConstraintKalman2007,genevaOpenVINSResearchPlatform2020}.
Storage is bounded by an $O(n_{\text{state}}^2)$ running factor plus one track's incoming rows, independently of track count.
The strategy is applicable to other memory-constrained MSCKF implementations.

As in OpenVINS~\cite{genevaOpenVINSResearchPlatform2020}, an optional per-update track cap limits the number processed, prioritizing the longest tracks.
A very small cap reduces the memory benefit of streaming across tracks, while streaming board-pose elimination still bounds the cross-frame accumulator.

\subsection{Rigid-Board Measurement Co-Design}
\label{sec:meas_model}

Our rigid-board model exploits the constellations' known inter-LED geometry and placement on a common level plane through a shared board pose.
We examine how retaining this structure affects nuisance dimension, geometric constraints, and processing cost.
The model also applies to other rigid fiducials with known point geometry placed on a common known plane.

\parhead{Rigid-board model.}
Each constellation board carries $N_\ell$ LEDs at known positions $\bm{\ell}_i^{B} \in \R^3$, $i = 0, \ldots, N_\ell{-}1$, expressed in a board-fixed frame $\{B\}$.
Placement on the known level plane restricts the board-to-global transformation to two translations and one yaw angle:
\begin{equation}
  \mathbf{p}_i^{G} = \mat{R}(\theta)\,\bm{\ell}_i^{B} + \mathbf{t},
  \label{eq:board_pose}
\end{equation}
where $\mathbf{t} = [t_x,\, t_y,\, 0]^\top$ and $\mat{R}(\theta)$ is a rotation about the vertical axis.
We choose the ground plane as $z=0$.
The board-pose nuisance parameter is $\bm{\beta} = (t_x, t_y, \theta) \in \R^3$.

At camera clone~$f$, the $i$-th LED produces a pixel observation $\meas_{f,i} \in \R^2$.
The measurement residual is
\begin{equation}
  \mathbf{r}_{f,i}
  = \meas_{f,i} - \pi\!\bigl(\mat{R}_{G}^{C_f}(\mathbf{p}_i^{G} - \mathbf{p}_{C_f}^{G})\bigr),
  \label{eq:residual}
\end{equation}
where $\pi(\cdot)$ denotes the camera projection and $(\mat{R}_{G}^{C_f}, \mathbf{p}_{C_f}^{G})$ is the clone pose stored in the state.

Stacking all $N_\ell$ LEDs in frame~$f$ and linearizing gives
\begin{equation}
  \underbrace{\mathbf{r}_f}_{2N_\ell \times 1}
  \;=\;
  \underbrace{\mat{H}_{\beta}}_{2N_\ell \times 3}\,\delta\bm{\beta}
  \;+\;
  \underbrace{\mat{H}_{x_f}}_{2N_\ell \times 6}\,\delta\state_{c_f}
  \;+\; \mathbf{n}_f,
  \label{eq:linearized}
\end{equation}
where $\mat{H}_{\beta}$ is the Jacobian with respect to the shared board pose (obtained by differentiating~\eqref{eq:board_pose}) and $\mat{H}_{x_f}$ is the Jacobian with respect to the clone pose of frame~$f$.

\subsection{Board-Pose Recovery and Nullspace Elimination}
\label{sec:board_tri}
\label{sec:board_null}

\parhead{Pose recovery.}
For each clone frame, IPPE~\cite{collinsInfinitesimalPlaneBasedPose2014} recovers a board-to-camera pose from the five LED observations; the frame with the lowest reprojection error seeds the initial board-to-global pose via the corresponding clone's state.
A Gauss-Newton refinement over the 3 planar DOF $(t_x, t_y, \theta)$ then jointly minimizes reprojection error across all clone frames, with optional Cauchy IRLS weighting of whole-board observations and $\chi^2$ gating to suppress outliers; the robust weights also whiten residual and Jacobian rows before nullspace projection.

We eliminate the shared board pose using two-level streaming QR, preserving inter-LED constraints.
After innovation gating, the resulting state constraints enter the measurement-factor fold~\eqref{eq:fold}.

\parhead{Per-frame elimination.}
Givens rotations first triangularize the board-pose Jacobian, then compress the board-free rows over the clone's six pose columns.
Partitioning $\mat{Q} = [\mat{Q}_1 \;\; \mat{Q}_2 \;\; \mat{Q}_3]$ into groups of 3, $q=\min(2N_\ell{-}3,6)$, and the remaining columns gives
\begin{equation}
  \begin{bmatrix}
    \mat{Q}_1^\top \mathbf{r}_f \\[2pt]
    \mat{Q}_2^\top \mathbf{r}_f \\[2pt]
    \mat{Q}_3^\top \mathbf{r}_f
  \end{bmatrix}
  =
  {\setlength{\arraycolsep}{1pt}
  \begin{bmatrix}
    \underbrace{\mat{R}_b}_{3 \times 3} & \mat{Q}_1^\top \mat{H}_{x_f} \\[2pt]
    \mat{0}   & \underbrace{\mat{T}_f}_{q \times 6} \\[2pt]
    \mat{0}   & \underbrace{\mat{0}}_{(2N_\ell - 3 - q) \times 6}
  \end{bmatrix}}
  \begin{bmatrix} \delta\bm{\beta} \\ \delta\state_{c_f} \end{bmatrix}
  +
  \begin{bmatrix}
    \mat{Q}_1^\top \mathbf{n}_f \\[2pt]
    \mat{Q}_2^\top \mathbf{n}_f \\[2pt]
    \mat{Q}_3^\top \mathbf{n}_f
  \end{bmatrix},
  \label{eq:qr1}
\end{equation}
where $\mat{R}_b$ is upper triangular and $\mat{T}_f$ is upper trapezoidal (triangular when $q=6$).
The top three \emph{board-coupled} rows are retained for cross-frame accumulation; the middle $q$ \emph{board-free} rows enter the shared fold~\eqref{eq:fold} for accepted tracks.
The remaining rows have zero Jacobian but retain residual energy used in innovation gating, with the full projected residual degrees of freedom.
This local compression assumes fixed camera intrinsics and extrinsics, with no additional calibration columns.

\parhead{Cross-frame accumulation.}
A three-row accumulator combines the board-coupled rows across frames, retaining their board factor and associated state Jacobian.
Stacking the next frame's three coupled rows below it and applying Givens rotations restores the triangular board block:
\begin{equation}
  \underbrace{\begin{bmatrix}
    \mat{R}_b^{\text{acc}} & \mat{H}_{x}^{\text{acc}} \\[2pt]
    \mat{R}_{b,f}          & \mat{Q}_1^\top \mat{H}_{x_f}
  \end{bmatrix}}_{6 \,\times\, (3 + 6N)}
  \;\xrightarrow{\text{Givens}}\;
  \begin{bmatrix}
    \mat{R}_b^{\text{acc}\prime} & \mat{H}_{x}^{\text{acc}\prime} \\[2pt]
    \mat{0}                      & \tilde{\mat{H}}_{x}
  \end{bmatrix}.
  \label{eq:qr2}
\end{equation}
The top three rows carry forward; the bottom three board-free rows enter the shared fold~\eqref{eq:fold} for accepted tracks.
After the final frame, discarding the accumulator eliminates the board nuisance pose.

For $N_\ell\geq5$, local compression limits the rows passed to the shared fold to $9F-3$, independent of board size, while measurement evaluation still scales with LED count.
Unlike OV\_plane's batch second-stage projection~\cite{chenMonocularVisualInertialOdometry2023}, our cross-frame elimination maintains a three-row accumulator.

\subsection{Sensor Compensation and Initialization}
\label{sec:robustness}

\parhead{Rolling-shutter compensation.}
The NanEyeC's rolling shutter captures different image rows at different times.
Following the use of capture-time camera poses in rolling-shutter VIO~\cite{liRealTimeMotionTracking2013}, we assign each board observation an effective capture time from its mean readout row and extrapolate the clone pose using frozen motion estimates, without augmenting the state.

\parhead{Initialization.} \label{sec:init}
We support \emph{ground-truth initialization} for controlled evaluation and \emph{hybrid initialization} using static IMU estimates of tilt and biases with visual estimates of yaw and metric height.
For on-drone deployment, \emph{pad initialization} uses a 3D-printed launch pad above a constellation (Sec.~\ref{sec:launch_pad}) to recover the initial pose from LED observations, gravity, and the known camera height; biases come from a static IMU window.
The pad defines the navigation frame, and stationary updates maintain the estimate before takeoff~\cite{genevaOpenVINSResearchPlatform2020}.

\subsection{Measurement-Model Comparison}
\label{sec:model_comparison}

We compare the rigid-board model with two independent-point models using the same LED observations and downstream streaming update.
The standard MSCKF model estimates each LED's three-dimensional position independently~\cite{mourikisMultiStateConstraintKalman2007}; the planar-point model fixes its height on the known plane and estimates only its horizontal coordinates.
Given camera poses, planar-point recovery requires non-grazing ray--plane intersections rather than translational parallax, isolating the information supplied by coplanarity.
Neither point model retains the known inter-LED geometry.

Table~\ref{tab:measurement_structure} compares track counts, nuisance DOF per track, and total rows supplied to the shared fold for the same LED observations.
The board model retains clone-local constraints because a single frame supplies more observations than shared nuisance DOF; generic independent-point elimination instead requires combining frames.
The row counts assume full-rank nuisance Jacobians and do not represent independent observable state directions.

\begin{table}[!t]
\centering
\caption{Processing one board's $N_\ell$ LEDs over $F$ frames.}
\label{tab:measurement_structure}
\small
\setlength{\tabcolsep}{3pt}
\begin{tabular}{@{}llll@{}}
\toprule
Model & Tracks & DOF/track & Total fold rows \\
\midrule
Rigid board & $1$ & $3$ & $qF+3(F-1)$ \\
Planar points & $N_\ell$ & $2$ & $N_\ell(2F-2)$ \\
Free points & $N_\ell$ & $3$ & $N_\ell(2F-3)$ \\
\bottomrule
\end{tabular}
\end{table}

\section{Experimental Methodology}
\label{sec:exp_method}

We combine controlled simulation, HITL end-to-end captures, and on-drone experiments to evaluate the proposed system.

\subsection{Simulation Protocol}
\label{sec:sim_datasets}

To isolate the effects of modeling assumptions, we generate trajectories in Webots~\cite{michelWebotsProfessionalMobile2004} using a quadrotor model extended from CrazySim~\cite{llanesCrazySimSoftwareLoop2024}, with the Crazyflie firmware's estimator and controller.
Figure-eight paths vary in lap period, amplitude, and altitude.
Seven trajectories span \SIrange{0.35}{0.99}{\meter\per\second} mean speed, up to \SI{19}{\degree} tilt, and \SIrange{0.7}{1.2}{\meter} altitude over a $4{\times}4$ constellation grid with \SI{0.61}{\meter} spacing.
We hold the nominal board layout and LED pattern fixed; optimizing these spatial arrangements is left to future work.
Synthetic camera and IMU measurements are generated at \SI{30}{\hertz} and \SI{400}{\hertz}, respectively, with configurable detection noise, inertial noise, and bias random walks.
All visible boards are reported with correct labels, isolating backend performance from detection and association failures.

We report absolute trajectory error after \SI{4}{DOF} (yaw and translation) alignment~\cite{zhangTutorialQuantitativeTrajectory2018}.
Reported errors are medians over all runs in a configuration, pooled across the seven equally represented trajectories.
Runs exceeding \SI{50}{\centi\meter} ATE remain in the sample and are also reported as a divergence rate; the median is less sensitive to extreme errors than the mean.
For each tested value of each swept parameter, we evaluate all seven trajectories with \num{50} noise and perturbation seeds per trajectory.

In detection-noise sweeps, the generated and estimator-assumed noise levels vary together.
Separate sweeps vary LED placement error, ground-plane height variation, camera principal-point and distortion error, clone-window length, the $\chi^2$ gate multiplier, and the per-update measurement budget.

\subsection{Hardware and Shared Testbed}
\label{sec:proto_hw}

For hand-held evaluation, a 3D-printed rig holds Vicon markers and a custom carrier PCB combining a Raspberry Pi Pico~2 with NanEyeC camera and ICM-45686 IMU evaluation boards.
This platform preserves the drone's sensing and compute architecture, using an RP2350 with external flash in place of the RP2354.
Both platforms use the same camera exposure and an 8-bit pixel threshold of 127 for event generation.

Experiments use a Vicon motion-capture arena with constellation boards arranged on a flat surface in $2{\times}2$ to $4{\times}4$ grids at approximately \SI{0.61}{\meter} spacing, matching the spacing used in simulation.
For hand-held captures, a LabJack T4 records camera frame-sync and Vicon timing signals for sub-millisecond alignment.

Camera intrinsics (equidistant model) are calibrated using an AprilGrid target~\cite{olsonAprilTagRobustFlexible2011}, camera--IMU extrinsics using Kalibr~\cite{furgaleUnifiedTemporalSpatial2013}, and the Vicon-to-sensor transform using hand--eye calibration~\cite{furrerEvaluationCombinedTimestamp2018}.

\subsection{HITL Evaluation}
\label{sec:hitl}

HITL evaluation runs the full frontend and estimator live on the microcontroller while the rig is moved through the motion-capture arena.
Previously recorded HITL experiments were used to tune the system.
Each live capture runs one measurement model, so comparisons on identical inputs use host replay of each recording through both models.
Paired replays use identical recorded frontend observations, initialization procedures, calibration, and per-update LED budgets.\footnote{The accompanying video (\url{https://youtu.be/6hQNdIcjHJE}) demonstrates the experimental setups and system operation.}

For HITL, we score live and replayed estimates after hybrid initialization against Vicon ground truth using SE(3) alignment~\cite{zhangTutorialQuantitativeTrajectory2018}.
We report position RMSE (ATE), orientation RMSE, and translational relative pose error (RPE) over \SI{10}{\meter} segments, expressed as RMSE normalized by segment length.
Path length and median ground-truth speed are computed over the scored interval.
Replay summaries average the per-capture errors with equal weight.

Device timing is measured during live execution.
Frontend (FE) means use the post-initialization portion of the first \num{1024} recorded per-frame timing samples; backend (BE) means and observed maxima are reported per capture.
Mean estimate latency measures time from the camera measurement timestamp to availability of the corrected estimate, subtracting the \SI{34.2}{\milli\second} frame period at \SI{29.2}{\hertz} from the measured gate-to-pose interval.

\label{sec:fe_gt}During live HITL experiments, surveyed board poses and Vicon measurements provide reference LED projections for evaluating detection recall and labeling.
Per-LED localization error is measured relative to each board's rigid pattern to suppress common errors in the ground-truth projection chain.

\subsection{On-Drone Evaluation}
\label{sec:on_drone}

We modify the Crazyflie firmware to communicate with the TinyCVIO deck over I\textsuperscript{2}C (Fig.~\ref{fig:sys_arch}).
Flights use Crazyswarm~\cite{preissCrazyswarm2017} with Vicon-corrected Crazyflie feedback to isolate estimator performance; TinyCVIO runs onboard, with its estimates logged over Crazyradio.
On-drone experiments use pad initialization (Sec.~\ref{sec:init}); the accompanying video shows the flight setup and operation.
The deck publishes poses over I\textsuperscript{2}C at \SI{50}{\hertz}, with Vicon ground truth at \SI{100}{\hertz}.
We score the airborne window (Vicon $z > \SI{0.6}{\meter}$) using the HITL alignment and error metrics, with a constant body-side mounting rotation fitted per run for orientation error.
A constant temporal offset is fitted per flight by minimizing airborne ATE.
Peak latency is the maximum recorded since boot across the three flights at each speed.
\label{sec:launch_pad}

\section{Results}
\label{sec:results}

We evaluate TinyCVIO in three stages: frontend detection and tracking on real hardware with Vicon ground truth, backend estimation in simulation, and the end-to-end system.

\subsection{Frontend Evaluation}
\label{sec:results_frontend}

We evaluate LED localization, labeling accuracy, and detection recall against the surveyed ground truth described in Sec.~\ref{sec:fe_gt}.

\begin{figure}[!t]
  \centering
  \includegraphics[width=0.99\columnwidth]{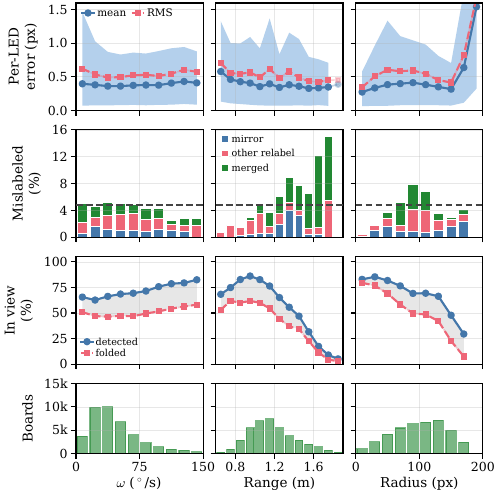}
  \caption{Frontend performance over six hand-held captures, binned by angular velocity, board distance, and image radius. Rows show localization error (mean, RMS, 5--95\% band), mislabeled detections by type (dashed: aggregate), in-view boards excluding the tracker's border exclusion zone that are detected (solid) and accepted by the backend (dashed), and board counts.}
  \label{fig:frontend_gt_eval}
\end{figure}

Detected constellations have an overall LED localization RMS error of \SI{0.6}{px} (Fig.~\ref{fig:frontend_gt_eval}).
Error varies little across the tested viewing distances and angular velocities.
Localization error also varies little with translational speed, which is not shown in Fig.~\ref{fig:frontend_gt_eval}.
Error increases near the image border, where viewing geometry and LED appearance become less favorable.

The frontend detects 82\% of in-view boards within \SI{1.4}{\meter} and inside \SI{140}{px} of the image center.
Being in view does not guarantee a resolvable constellation: LEDs can merge or become difficult to distinguish as distance and viewing angle change.
Among missed boards, 91\% remain bright, indicating that brightness alone does not ensure detection.

The label-continuity guard reduces mislabeled detections from 22.8\% to 4.9\% and increases the share entering accepted backend updates from 43\% to 52\%.
Errors typically begin as mid-track label flips following merged-LED observations.
Suppressing these flips preserves more observations for accepted backend updates.

\subsection{Backend Evaluation}
\label{sec:results_backend}

\begin{figure}[!t]
  \centering
  \includegraphics[width=0.99\columnwidth]{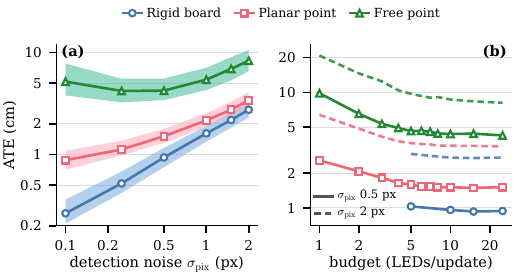}
  \caption{Median ATE over \num{350} runs per point, all axes log-scaled: (a) detection noise, with interquartile bands; (b) per-update LED budget at \SI{0.5}{px} (solid) and \SI{2}{px} (dashed) noise. Board budgets increase in steps of five LEDs. Free points diverge in 3--15\% of runs; neither constrained model diverges.}
  \label{fig:backend_models}
\end{figure}

Unless varied in a sweep, simulations use six clones, a $\chi^2$ gate multiplier of \num{1}, Cauchy weighting, and no per-update measurement cap.
All three models receive identical observations, isolating the effects of their geometric constraints.

\parhead{Measurement model.}
The rigid-board model achieves the lowest median ATE across the tested detection-noise levels (Fig.~\ref{fig:backend_models}a).
Its advantage over planar points demonstrates the benefit of retaining inter-LED geometry beyond the ground-plane constraint.
As detection noise decreases, rigid-board accuracy continues to improve, while free-point accuracy reaches a floor.
The accuracy advantage depends on detection noise: rigid-board ATE is eight times lower than free-point ATE at \SI{0.25}{px} noise and three times lower at \SI{2}{px}.
Increasing the clone window from six to sixteen frames reduces ATE by 35\% for rigid boards, compared with 25\% and 17\% for planar and free points, respectively.
Longer clone windows increase computational and memory requirements; fold workspace grows from \SI{41.3}{} to \SI{56.1}{\kilo\byte} over this range.

\parhead{Model assumptions.}
Additional sweeps, not shown in Fig.~\ref{fig:backend_models}, reveal the cost of imposing geometric constraints.
A \SI{2}{\milli\meter} LED placement error increases rigid-board ATE by 66\%, while changing either point model's error by less than 3\%.
Ground-plane height variation affects both plane-constrained models more than free points.
A \SI{2}{px} principal-point error changes ATE by at most 3\% for all three models, whereas a 1.5\% distortion error increases ATE by 17--27\%.
Distortion error degrades all three models without changing their accuracy ranking.

\parhead{Gate and weighting.}
With uniform weighting, a $\chi^2$ gate multiplier of \num{0.5} causes the rigid-board model to diverge in 82\% of runs.
With Cauchy weighting, the rigid-board model shows no observed divergence across tested gate multipliers of \numrange{0.5}{20}.

\parhead{Measurement budget.}
We compare per-update measurement budgets in LEDs, with five LEDs per board (Fig.~\ref{fig:backend_models}b).
Limiting updates to one board reduces fold workspace to one-quarter of the uncapped requirement while increasing ATE by approximately 10\%.
At matching detection noise, this configuration remains more accurate than either independent-point model at any tested budget.
Across the tested detection-noise levels, its ATE stays 9--11\% above the uncapped result, whereas free points require larger budgets as noise increases.

\begingroup
\setlength{\intextsep}{6pt}
\begin{table}[H]
\centering
\caption{Live HITL estimation accuracy and timing.}
\label{tab:hitl-results}
\small
\setlength{\tabcolsep}{3pt}
\begin{tabular*}{\columnwidth}{@{\extracolsep{\fill}}lrrr@{\hspace{10pt}}rrr@{}}
\toprule
 & \multicolumn{3}{c}{Rigid board} & \multicolumn{3}{c}{Planar point} \\
\cmidrule(lr){2-4}\cmidrule(l){5-7}
Capture & B1 & B2 & B3 & P1 & P2 & P3 \\
\midrule
Path (m) & 39.8 & 45.0 & 41.8 & 52.3 & 53.1 & 30.5 \\
Median speed (m/s) & 0.77 & 0.87 & 0.81 & 0.98 & 1.04 & 0.47 \\
\midrule
ATE (cm) & 1.92 & 3.43 & 1.75 & 2.72 & 3.61 & 2.81 \\
Orientation (${}^\circ$) & 1.13 & 1.35 & 1.63 & 1.84 & 2.39 & 1.17 \\
RPE$_{10\,\mathrm{m}}$ (\%) & 0.29 & 0.38 & 0.26 & 0.37 & 0.43 & 0.39 \\
\midrule
FE mean (ms) & 5.0 & 4.9 & 5.4 & 5.1 & 5.0 & 4.1 \\
BE mean (ms) & 14.2 & 13.5 & 14.1 & 15.5 & 15.6 & 17.5 \\
BE max. (ms) & 25.7 & 24.8 & 24.7 & 28.8 & 30.0 & 29.4 \\
Mean est. latency (ms) & 18.7 & 17.8 & 18.7 & 20.0 & 20.2 & 21.4 \\
\bottomrule
\end{tabular*}
\end{table}
\endgroup

\subsection{HITL Evaluation}
\label{sec:results_integrated}

Both models use six clones, a $\chi^2$ gate multiplier of 1, Cauchy weighting, the label-continuity guard, and the equidistant camera calibration.
Each update processes at most two board tracks or ten individual LED tracks, corresponding to the same LED budget.
Table~\ref{tab:hitl-results} reports three live captures with the rigid-board model (B1--B3) and three with planar points (P1--P3), all using the configuration above and each spanning approximately \SI{51.5}{\second} of evaluated hand-held motion.
Table~\ref{tab:hitl-replay} reports mean errors across 19 paired capture replays, with per-capture median ground-truth speeds ranging from \SIrange{0.47}{1.04}{\meter\per\second}.

Averaged across these live captures, mean backend processing time is approximately 14\% lower for the rigid-board model than for planar points.
This reduction comes from processing each board as a shared track in the measurement model, rather than recovering and processing its LEDs independently.
On identical recorded inputs, it reduces mean ATE by 27\%, with lower ATE on 18 of 19 captures and lower mean orientation and relative translation errors.

\begingroup
\setlength{\intextsep}{6pt}
\begin{table}[H]
\centering
\caption{Mean errors over 19 paired HITL capture replays.}
\label{tab:hitl-replay}
\small
\setlength{\tabcolsep}{4pt}
\begin{tabular}{@{}lrrr@{}}
\toprule
Model & ATE (cm) & Ori. (${}^\circ$) & RPE$_{10\,\mathrm{m}}$ (\%) \\
\midrule
Rigid board & \textbf{2.44} & \textbf{1.91} & \textbf{0.34} \\
Planar point & 3.32 & 2.38 & 0.46 \\
\bottomrule
\end{tabular}
\end{table}
\endgroup

\subsection{On-Drone Demonstration}
\label{sec:results_drone}

\begin{table}[!t]
\centering
\caption{Flight results across three trials per speed.}
\label{tab:flight}
\small
\setlength{\tabcolsep}{3pt}
\begin{tabular*}{\columnwidth}{@{\extracolsep{\fill}}lrrr@{}}
\toprule
Speed (m/s) & 0.34 & 0.41 & 0.50 \\
\midrule
ATE (cm)                & 3.65 $\pm$ 0.43 & 3.48 $\pm$ 0.34 & 3.71 $\pm$ 0.79 \\
Ori. (${}^\circ$)       & 1.44 $\pm$ 0.31 & 1.31 $\pm$ 0.68 & 1.20 $\pm$ 0.26 \\
RPE$_{10\,\mathrm{m}}$ (\%) & 0.57 $\pm$ 0.06 & 0.50 $\pm$ 0.10 & 0.60 $\pm$ 0.15 \\
\midrule
Mean est. latency (ms)       & 15.7 $\pm$ 0.1  & 15.7 $\pm$ 0.1  & 16.3 $\pm$ 0.1 \\
Peak est. latency (ms)         & 30.7            & 31.9            & 31.2 \\
\bottomrule
\end{tabular*}
\end{table}

\begin{figure}[!t]
  \centering
  \includegraphics[width=0.99\columnwidth]{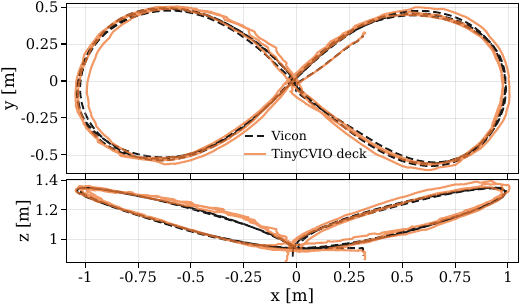}
  \caption{Estimated and Vicon trajectories for one flight in the fastest speed group, after SE(3) alignment.}
  \label{fig:flight-traj}
\end{figure}

We evaluate onboard estimation with the rigid-board model along a figure-eight trajectory with two \SI{0.40}{\meter} vertical excursions per lap, flown three times each at mean ground-truth speeds of 0.34, 0.41, and \SI{0.50}{\meter\per\second}.
For flight, we increase the $\chi^2$ gating threshold multiplier to 5 and enable the IMU's onboard low-pass filter, extensively tested beforehand, to attenuate motor vibration.
Table~\ref{tab:flight} reports mean $\pm$ standard deviation over three flights per speed and peak estimate latency, and Fig.~\ref{fig:flight-traj} shows one flight from the fastest group.

Position accuracy remains consistent across the tested speeds, with greater variability in the fastest group.
Estimate latency remains stable, demonstrating real-time onboard estimation during repeated flights.
Flight execution has lower instrumentation overhead than HITL, which records additional data for replay and diagnostics.

\section{Conclusion}
\label{sec:conclusion}

TinyCVIO co-designs miniature sensing, visual processing, and visual-inertial estimation around LED constellations on a commodity microcontroller.
Retaining known constellation geometry improves estimation accuracy, while streaming QR bounds workspace for a fixed state size.
Simulation, paired replays, and live hardware experiments demonstrate the benefits of this design, culminating in real-time onboard estimation during repeated Crazyflie flights.

\section*{ACKNOWLEDGMENT}
This work was supported in part by a Cornell College of Engineering SPROUT award.
Claude (Anthropic) was used as a coding and debugging aid and to assist in the manuscript editing process for clarity.

\bibliographystyle{IEEEtran}
\bibliography{references-compact}

\end{document}